%% file: main.tex
\documentclass[10pt,twocolumn,letterpaper]{article}

\usepackage{cvpr}
\usepackage{times}
\usepackage{epsfig}
\usepackage{graphicx}
\usepackage{amsmath}
\usepackage{amssymb}
\usepackage{comment}
\usepackage{multirow}
\usepackage{xcolor}
\usepackage{caption} 
\usepackage{physics} 
\usepackage[pagebackref=true,breaklinks=true,letterpaper=true,colorlinks,bookmarks=false]{hyperref}

\usepackage[accsupp]{axessibility}  
\DeclareUnicodeCharacter{1EEF}{\u{u}}
\DeclareUnicodeCharacter{1EC5}{\~{\^e}}

\cvprfinalcopy 

\definecolor{lightgray}{RGB}{220,220,220}
\definecolor{darkblue}{RGB}{0,0,127}
\definecolor{darkgreen}{RGB}{0,127,0}
\definecolor{darkred}{RGB}{200,0,0}

\ifcvprfinal\fi

\usepackage[affil-it]{authblk}

\usepackage{enumitem}

\title{The 8th AI City Challenge}

\begin{document}

\pagenumbering{gobble}

\author{
Shuo Wang$^1$ \hspace{0.8cm}
David C. Anastasiu$^2$ \hspace{0.8cm}
Zheng Tang$^1$ \hspace{0.8cm}
Ming-Ching Chang$^3$ \\
Yue Yao$^4$ \hspace{0.8cm}
Liang Zheng$^4$ \hspace{0.8cm}
Mohammed Shaiqur Rahman$^5$ \hspace{0.8cm}
Meenakshi S. Arya$^5$ \\
Anuj Sharma$^5$ \hspace{0.8cm}
Pranamesh Chakraborty$^6$ \hspace{0.8cm}
Sanjita Prajapati$^6$ \hspace{0.8cm}
Quan Kong$^7$ \\
Norimasa Kobori$^7$ \hspace{0.6cm}
Munkhjargal Gochoo$^{8,11}$ \hspace{0.6cm}
Munkh-Erdene Otgonbold$^{8,11}$ \hspace{0.6cm}
Fady Alnajjar$^8$ \\
Ganzorig Batnasan$^8$ \hspace{0.8cm}
Ping-Yang Chen$^9$ \hspace{0.8cm}
Jun-Wei Hsieh$^9$ \hspace{0.8cm}
Xunlei Wu$^1$ \\
Sameer Satish Pusegaonkar$^1$ \hspace{0.8cm}
Yizhou Wang$^1$ \hspace{0.8cm}
Sujit Biswas$^1$ \hspace{0.8cm}
Rama Chellappa$^{10}$
} 
\affil{ 
$^1$ NVIDIA Corporation, CA, USA \hspace{0.8cm} 
$^2$ Santa Clara University, CA, USA \\ 
$^3$ University at Albany, SUNY, NY, USA \hspace{0.8cm} 
$^4$ Australian National University, Australia \\
$^5$ Iowa State University, IA, USA \hspace{0.8cm}
$^6$ Indian Institute of Technology Kanpur, India \\
$^7$ Woven by Toyota, Japan \hspace{0.8cm}
$^8$ United Arab Emirates University, UAE \\
$^9$ National Yang-Ming Chiao-Tung University, Taiwan \hspace{0.8cm}
$^{10}$ Johns Hopkins University, MD, USA \\
$^{11}$ Emirates Center for Mobility Research, UAE
}

\maketitle

\input{0Abstract.tex}

\input{1Introduction.tex}

\input{2ChallengeSetup.tex}

\input{3Datasets.tex}

\input{4Evaluation.tex}

\input{5SubmissionResults.tex}

\input{6Conclusion.tex}

\input{7Acknowledgement.tex}


{\small
\bibliographystyle{ieee_fullname}
\bibliography{aicity24, aicity23, aicity22, aicity21, aicity20, aicity19, aicity18, aicity17}
}

\end{document}

%% file: 0Abstract.tex
\begin{abstract}
The eighth AI City Challenge highlighted the convergence of computer vision and artificial intelligence in areas like retail, warehouse settings, and Intelligent Traffic Systems (ITS), presenting significant research opportunities. The 2024 edition featured five tracks, attracting unprecedented interest from 726 teams in 47 countries and regions. Track 1 dealt with multi-target multi-camera (MTMC) people tracking, highlighting significant enhancements in camera count, character number, 3D annotation, and camera matrices, alongside new rules for 3D tracking and online tracking algorithm encouragement. Track 2 introduced dense video captioning for traffic safety, focusing on pedestrian accidents using multi-camera feeds to improve insights for insurance and prevention. Track 3 required teams to classify driver actions in a naturalistic driving analysis. Track 4 explored fish-eye camera analytics using the FishEye8K dataset. Track 5 focused on motorcycle helmet rule violation detection. The challenge utilized two leaderboards to showcase methods, with participants setting new benchmarks, some surpassing existing state-of-the-art achievements.

\end{abstract}

%% file: 1Introduction.tex
\section{Introduction}

The AI City Challenge, showcased at CVPR 2024, leverages artificial intelligence to boost operational efficiency in physical settings, including retail and warehouse environments, as well as Intelligent Traffic Systems (ITS). This initiative aims to derive actionable insights from sensor data, such as camera feeds, to enhance traffic safety and optimize transportation outcomes. The focus for this year centers on two pivotal areas poised for substantial impact: retail business operations and ITS, where the application of AI promises to usher in significant advancements.

Emphasizing practical, scalable applications, the Challenge called for original contributions across several critical domains: multi-camera people tracking, traffic safety analysis, naturalistic driving action recognition, fish-eye camera road object detection, and motorcycle helmet rule compliance. These areas represent the cutting edge in employing computer vision, natural language processing, and deep learning to bolster safety and intelligence within various environments. The 8th edition of the Challenge marks a milestone with the introduction of novel tasks and significant enhancements to datasets, including dense video captioning for traffic safety, fish-eye camera analytics with the FishEye8K dataset~\cite{UAEU23FishEye8K}, and substantial updates in multi-camera people tracking, featuring extensive increases in camera and character counts, alongside new rules and technologies like 3D tracking.

The five tracks of the AI City Challenge 2024 are summarized as follows:

\begin{itemize}[leftmargin=12pt] 

\item \textbf{Multi-target multi-camera (MTMC) people tracking:}
Participants in the challenge were supplied with videos from diverse synthetic indoor environments, with the main goal being to track individuals across the fields of view of different cameras. Camera matrices were made available to facilitate the inference of 3D positions. A preference was given to the use of online tracking algorithms, with bonuses awarded to teams utilizing these methods in determining the winners.

\item \textbf{Traffic safety description and analysis:}
This task focuses on the detailed video captioning of traffic safety scenarios, particularly involving pedestrian incidents, using the Woven Traffic Safety (WTS) dataset \cite{WTS2024}. Participants need to describe the moments leading up to the incidents and the general scene, noting relevant details about the context, attention to safety, location, and the behavior of both pedestrians and vehicles. This task offers an in-depth opportunity to analyze traffic safety scenarios.

\item \textbf{Naturalistic driving action recognition:}
In this competition track, teams were tasked with classifying 16 types of distracted driving behaviors such as texting, making phone calls, and reaching back. The Synthetic Distracted Driving (SynDD2) dataset \cite{rahman2023synthetic}, collected using three cameras inside a stationary vehicle, was employed. This year, the dataset size increased to 84 instances, up from 30 the previous year.

\item \textbf{Road object detection in fisheye cameras:}
Fisheye lenses are favored for their wide, natural, and omnidirectional field of view, providing coverage that traditional narrow-view cameras cannot. In traffic monitoring, fisheye cameras reduce the need for multiple cameras at street intersections but introduce challenges in image distortion. Teams were tasked with detecting five types of road objects (pedestrians, bikes, cars, trucks, and buses) in images from fisheye cameras.

\item \textbf{Detecting violation of helmet rule for motorcyclists:}
Teams were required to determine whether motorcyclists were wearing helmets—a safety measure mandated by laws in many countries. Automated detection of helmet non-compliance can significantly enhance the enforcement of traffic safety regulations.

\end{itemize}

The AI City Challenge continued to attract considerable interest and participation in its latest edition, similar to previous years. From the announcement of the challenge tracks in late January, participation requests surged to 726 teams, marking a 43\% increase from the 508 teams in 2023, with representation from 47 countries and regions globally. The distribution of team participation across the five challenge tracks was as follows: tracks 1 through 5 saw 421, 359, 349, 403, and 419 teams, respectively. Notably, this year, 209 teams registered for the evaluation system, a significant increase from the previous year's 159. The number of submissions for tracks 1, 2, 3, 4, and 5 were 17, 15, 16, 70, and 60, respectively.

This paper provides a comprehensive overview of the preparation and outcomes of the 8th AI City Challenge. Subsequent sections detail the setup of the challenge ($\S$\ref{sec:challenge:setup}), preparation of the challenge data ($\S$\ref{sec:dataset}), evaluation methodology ($\S$\ref{sec:eval}), analysis of the submitted results ($\S$\ref{sec:results}), and discuss the implications of the findings and directions for future research ($\S$\ref{sec:conclusion}).

%% file: 2ChallengeSetup.tex
\section{Challenge Setup}
\label{sec:challenge:setup}

The 8th AI City Challenge made its training and validation datasets available to participants on January 22, 2024, and subsequently released the test sets with the evaluation server's launch on February 19, 2024. The deadline for all challenge track submissions was set for March 25, 2024. Competitors aiming for prizes were mandated to open-source their code for verification purposes and ensure their code repositories were publicly accessible. This requirement stems from the expectation that winning teams would significantly contribute to the community and expand the existing knowledge base. Additionally, it was imperative for the results showcased on the leaderboards to be reproducible independently of any private data.

\textbf{Track 1: MTMC People Tracking.}
Teams in the challenge are required to track individuals across an array of cameras using a significantly expanded synthetic dataset. The dataset's scale has been notably increased: the camera count has surged from 129 to roughly 1,300, and the number of tracked individuals has grown from 156 to about 3,400. To assist teams, 3D annotations and camera matrices are provided. The evaluation metric has been updated to the Higher Order Tracking Accuracy (HOTA), which now considers 3D distances, offering a more detailed assessment of tracking precision. A new feature of this challenge encourages the adoption of online tracking, where algorithms predict current frame results based solely on past frame data. Submissions utilizing online tracking methods will benefit from a 10\% bonus to their HOTA score, a factor that could be decisive in close competitions for the top positions.

\textbf{Track 2: Traffic Safety Description and Analysis.}
In this challenge, teams will analyze video segments of traffic events, providing two detailed captions for each segment that describe the behavior of pedestrians and vehicles before and during accidents, as well as during normal traffic conditions. The descriptions should focus on location, attention, behavior, and context. The provided ground truth file includes captions and bounding box information for target instances. Evaluation will be based on several metrics assessing the accuracy of the predicted descriptions relative to the ground truth.

\textbf{Track 3: Naturalistic Driving Action Recognition.}
This track involves analyzing approximately 76 hours of video collected from 84 different drivers. Each team must submit a text file detailing one identified driving activity per line, including the start and end times, along with corresponding video file information. Performance is evaluated based on the accuracy of activity identification, specifically the average activity overlap score. The team with the highest score will be declared the winner.

\textbf{Track 4: Road Object Detection in Fisheye Cameras.}
Teams are tasked with detecting road objects (pedestrians, bikes, cars, trucks, and buses) in images from fisheye cameras. The challenge involves the FishEye1K\_eval test dataset, which consists of 1,000 images, and the FishEye8K training dataset, which includes 8,000 images. Both datasets were sourced from fisheye traffic surveillance cameras operated by the Hsinchu City Police Department in Taiwan.

\textbf{Track 5: Detecting Violation of Helmet Rule for Motorcyclists.}
Participants in this track are required to detect whether motorcycle drivers and passengers are wearing helmets, using traffic camera footage from an Indian city. The challenge categorizes drivers and passengers as separate entities and includes complex real-world scenarios characterized by poor visibility conditions, such as low light or fog, high traffic congestion at intersections, {\em etc.}

%% file: 3Datasets.tex
\section{Datasets}
\label{sec:dataset}

The datasets for the five challenge tracks of the 8th AI City Challenge are introduced as follows.


\subsection{The MTMC People Tracking Dataset}

\begin{figure}[t]
\centering
\includegraphics[width=0.47\textwidth]{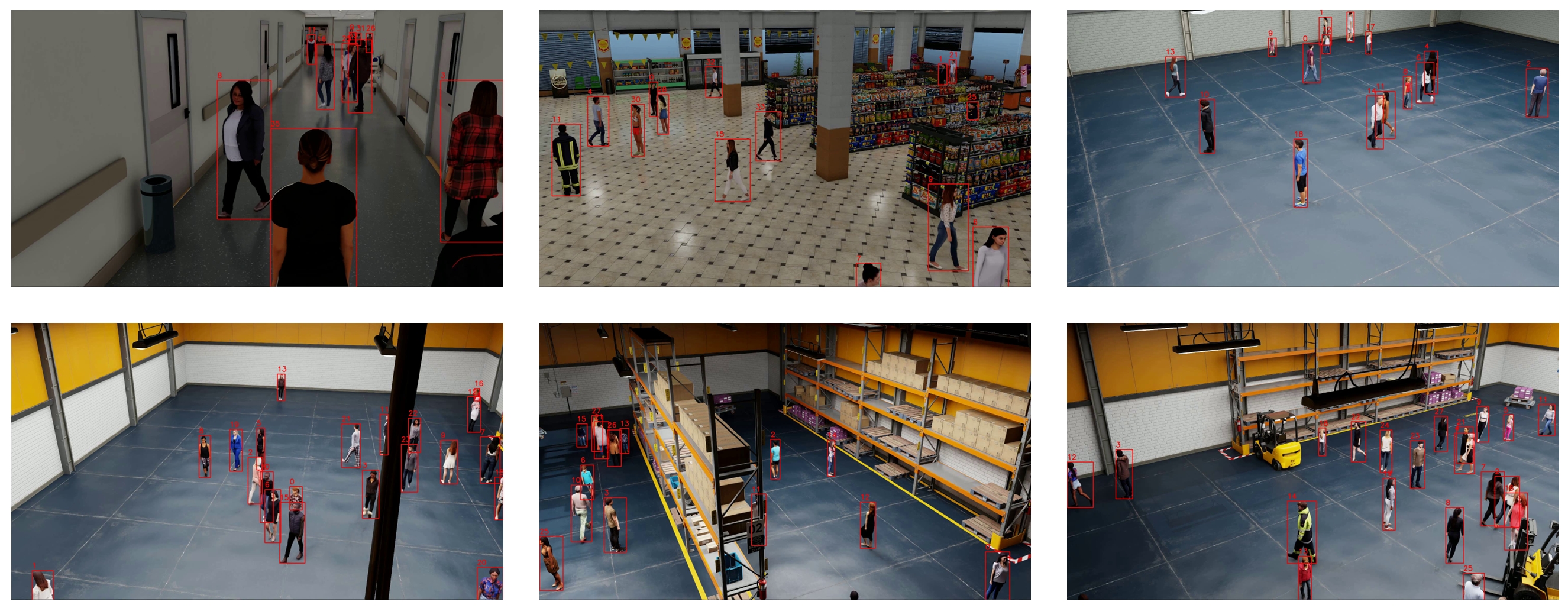}
\caption{The MTMC people tracking dataset for Track 1 contains 90 subsets from 6 synthetic environments. The figure contains sampled frames with plotted labels from the 6 environments.}
\label{fig:mtmc_people_tracking}
\end{figure}

The MTMC people tracking dataset, a comprehensive benchmark consisting of six different synthetic environments, was developed using the NVIDIA Omniverse Platform (see Figure~\ref{fig:mtmc_people_tracking}). This dataset encompasses 90 subsets—40 for training, 20 for validation, and 30 for testing—featuring 953 cameras, 2,491 people, and over 100 million bounding boxes, marking a significant expansion from the previous year's 22 scenes, 129 cameras, 156 people, and 8 million bounding boxes. With a total video length of 212 hours, presented in high-definition (1080p) at 30 frames per second, this benchmark surpasses its predecessors not only in scale but also in providing annotations of 3D locations and camera matrices, enabling 3D space MTMC tracking.

The ``Omniverse Replicator'' framework, instrumental for character labeling and synthetic data generation, annotates the camera-rendered output and formats it for learning utilization. The ``omni.anim.people'' extension is used for simulating human behaviors realistically in various synthetic environments. A workflow scheduling script, designed to operate automatically based on specific configurations, facilitated the efficient generation of this extensive MTMC dataset.

\subsection{The Woven Traffic Safety Dataset}

\begin{figure}[t]
\centering
\includegraphics[scale=0.25]{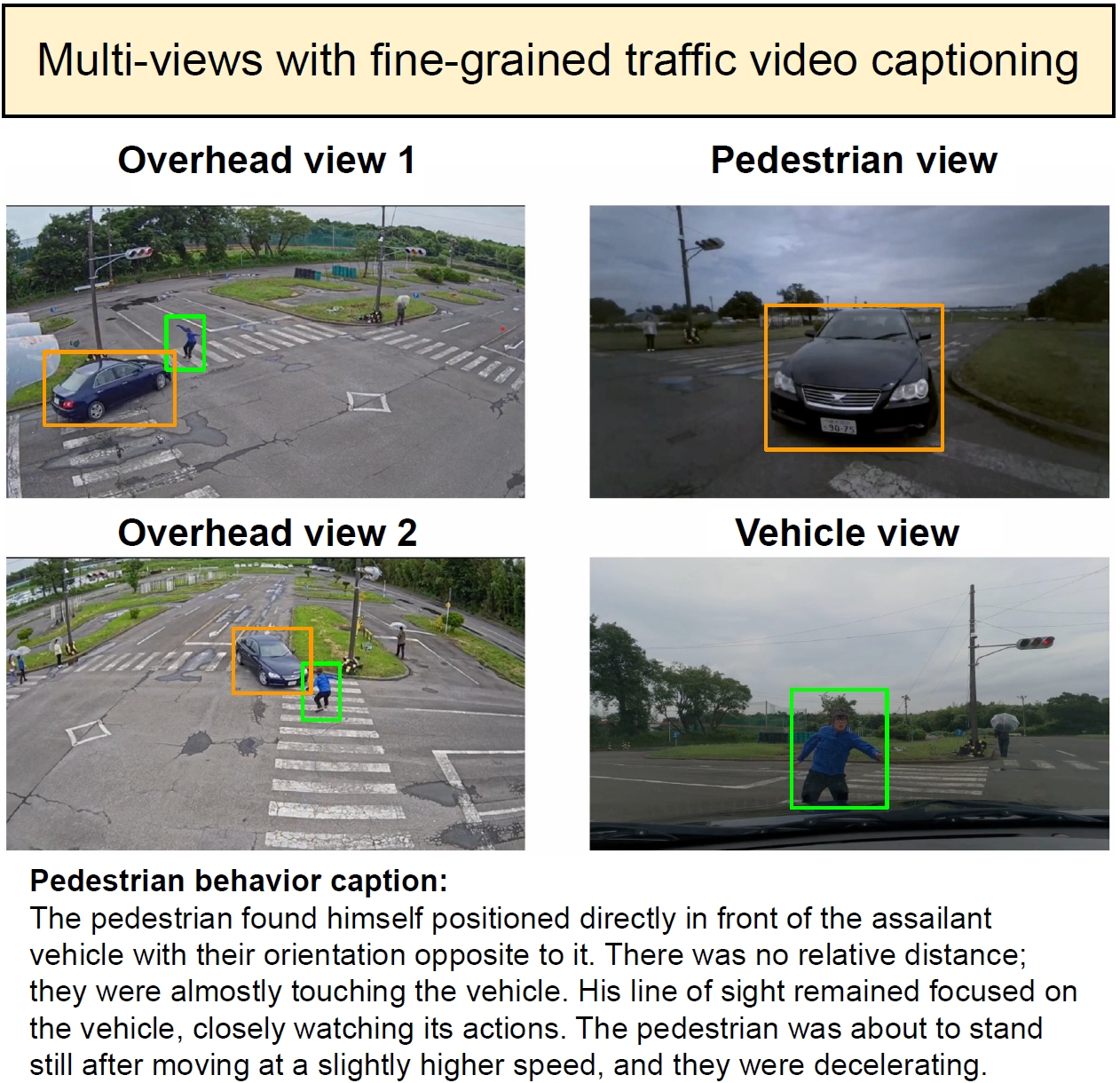}
   \caption{Overview of the WTS dataset for Track 2, providing multi-view videos with fine-grained captions focused on pedestrian perspectives.}
\label{fig:wts_intro}
\end{figure}

The Woven Traffic Safety (WTS) dataset~\cite{WTS2024} comprises train and validation sets with 810 multi-view videos of staged traffic scenarios, as shown in Figure~\ref{fig:wts_intro}. Each scenario is segmented into approximately 5 phases: \textit{pre-recognition, recognition, judgment, action,} and \textit{avoidance}, with each segment featuring 2 detailed captions. These captions are derived from a manual checklist of over 180 items related to the environmental context, attributes, position, action, and attention of pedestrians and vehicles. The items were processed using GPT-3.5~\cite{gpt3.5} to generate natural sentences that were then manually verified to establish the final ground truth. Each caption averages about 58.7 words in length. Additionally, the dataset includes about 3.4K fine-grained caption annotations from the BDD100K~\cite{Yu_2020_CVPR}, selected to enhance the generalizability of the models trained on this dataset.

\subsection{The {\bf \textit{SynDD2}} Dataset}

\textit{SynDD2}~\cite{rahman2023synthetic} includes 504 video clips in the training set and 90 videos in the test set, all recorded at 30 frames per second and at a resolution of 1920×1080. The videos are manually synchronized across three camera views~\cite{9857426} and are approximately 9 minutes in length. Each video showcases 16 distracted driving activities performed in random order and for varying durations, sometimes with an appearance block like a hat or sunglasses. Drivers contributed six videos each: three with an appearance block and three without.

\subsection{The {\bf \textit{FishEye8K}} and {\bf \textit{FishEye1Keval}} Datasets}

\begin{figure}[t]
\centering
\includegraphics[width=0.47\textwidth]{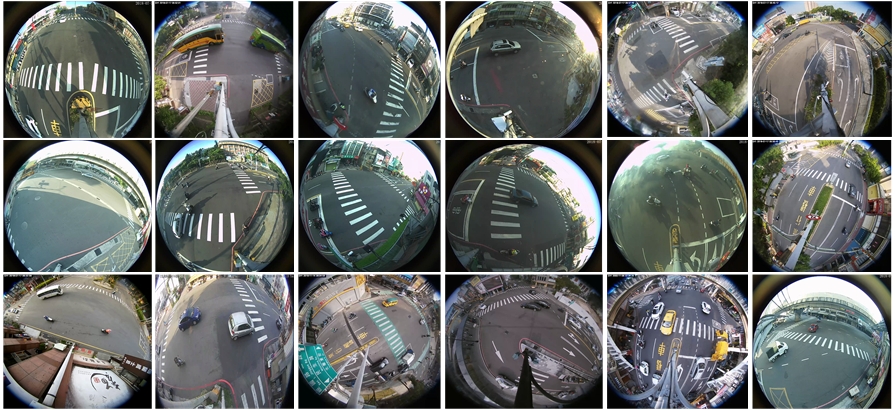}
\caption{Sample images from each of the 18 cameras with wide-angle fisheye views for Track 4.}
\label{fig:fisheye8K}
\end{figure}

The FishEye8K benchmark dataset, published in~\cite{Gochoo_2023_CVPR}, serves as both the training and validation sets, with 5,288 and 2,712 images respectively, featuring resolutions of 1080×1080 and 1280×1280. These sets contain a total of 157K annotated bounding boxes across five road object classes (Bus, Bike, Car, Pedestrian, Truck). The dataset was compiled from 35 fisheye videos recorded at 60 FPS using 20 traffic surveillance cameras in Hsinchu City, Taiwan. The FishEye1K\_eval test dataset, comprising 1,000 images, was extracted from 11 camera videos not used in the FishEye8K dataset. Dataset labels are available in XML (PASCAL VOC), JSON (COCO), and TXT (YOLO) formats.

\subsection{The Bike Helmet Violation Detection Dataset}

This dataset includes 100 videos each for the training and testing phases, recorded at 10 FPS and 1080p resolution from various locations in an Indian city. All pedestrian faces and vehicle license plates were redacted. The dataset features 9 object classes, including motorbike, \textit{DHelmet} (driver with helmet), \textit{DNoHelmet} (driver without helmet), \textit{P1Helmet} (first passenger with helmet), \textit{P1NoHelmet} (first passenger without helmet), \textit{P2Helmet} (second passenger with helmet), \textit{P2NoHelmet} (second passenger without helmet), \textit{P0Helmet} (child in front with helmet), and \textit{P0NoHelmet} (child in front without helmet). Bounding boxes have a minimum size of 40 pixels, similar to the KITTI dataset~\cite{geiger2012we}, and an object must be at least 40\% visible to be annotated. This year, the dataset has been enhanced to include more challenging scenarios such as congested traffic conditions and zoomed-in traffic camera views, akin to those used in traffic violation detection systems.

%% file: 4Evaluation.tex
\section{Evaluation Methodology}
\label{sec:eval}

As in previous AI City Challenges~\cite{Naphade18AIC18, Naphade19AIC19, Naphade20AIC20, Naphade21AIC21, Naphade22AIC22, Naphade23AIC23}, we employed an \textbf{online evaluation system} allowing teams to submit multiple solutions to each problem and automatically evaluated the performance in real time. The results were shared with the submitting team and other participants. An anonymized leaderboard displayed the top three results for each track to encourage ongoing improvement. Teams were limited to five submissions per day and 20--40 submissions per track overall, with submissions containing errors exempt from these limits. Initially, results were calculated using a random 50\% subset of the test set to prevent overfitting, with full test set scores revealed post-competition.

Teams competing for prizes were prohibited from using private data or manual labeling on the \textbf{Public} leaderboard, while others could submit to a separate \textbf{General} leaderboard.

\subsection{Track 1 Evaluation}

Contrary to our 2023 Challenge~\cite{Naphade23AIC23}, which used the IDF1 metric, this year we adopted the Higher Order Tracking Accuracy (HOTA) scores~\cite{luiten2020IJCV} for evaluation. HOTA is computed on the 3D locations of objects, with repetitive data points removed across cameras for the same frame. Euclidean distances between predicted and ground truth 3D locations are converted to similarity scores using a zero-distance parameter; scores are zero for distances over 2 meters. These scores contribute to the calculation of localization accuracy (LocA), detection accuracy (DetA), and association accuracy (AssA) using the TrackEval library \cite{luiten2020trackeval}.

\[
\text{HOTA}_{\alpha} = \sqrt{\text{DetA}_{\alpha} \cdot \text{AssA}_{\alpha}}, \label{eq:HOTA-partial}
\]
\[
\text{HOTA} = \int_{0}^{1} \text{HOTA}_{\alpha} \, d\alpha, \label{eq:Track1-HOTA}
\]
where $\alpha$ is the localization intersection-over-union (IOU) threshold, varying in 0.05 increments from 0 to 1. Submissions employing online tracking technologies receive a 10\% bonus to their HOTA scores.

\subsection{Track 2 Evaluation} 
\label{sec:track2:eval}

Teams are ranked based on averaged accuracy against the ground truth using multiple metrics across all scenarios from both the staged and BDD parts. Four metrics are averaged: BLEU-4 \cite{papineni-etal-2002-bleu}, METEOR \cite{banerjee-lavie-2005-meteor}, ROUGE-L \cite{lin-2004-rouge}, and CIDEr \cite{oliveira-dos-santos-etal-2021-cider}. Each video segment includes two captions, one for pedestrians and one for vehicles. To eliminate sample number bias between the staged and BDD parts, scores for each are calculated separately and then averaged to determine the final ranking.

\subsection{Track 3 Evaluation}
\label{sec:track1:eval}

The evaluation criteria for this track remain unchanged from last year~\cite{Naphade23AIC23}. Performance is measured by the average activity overlap score, calculated as follows: Given a ground-truth activity $g$ with start and end times $gs$ and $ge$, the closest predicted activity $p$ must match the class of $g$ and maximize the overlap score $os$ within a time window defined by $gs \pm 10s$ and $ge \pm 10s$. The overlap score is defined as:
\[
os(p,g) = \frac{\max(\min(ge,pe)-\max(gs,ps), 0)}{\max(ge,pe)-\min(gs,ps)}.
\]
All activities are processed in the order of their start times, and any unmatched activities receive a score of zero. The final score is the average of all overlap scores.

\subsection{Track 4 Evaluation}
\label{sec:track4:eval}

Track 4's evaluation is based on the \textit{F}1 score, defined as the harmonic mean of precision and recall:
\[
F1 = \frac{2 \times \text{Precision} \times \text{Recall}}{\text{Precision} + \text{Recall}}.
\]

\subsection{Track 5 Evaluation}
\label{sec:track5:eval}

Evaluation for Track 5 is based on mean Average Precision (mAP) across all test video frames, as defined in the PASCAL VOC 2012 competition~\cite{pascal-voc-2012}. The mAP score calculates the average of precision scores (area under the Precision-Recall curve) for all object classes. Bounding boxes smaller than 40 pixels or overlapping with redacted regions are excluded to prevent penalization errors.

%% file: 5SubmissionResults.tex
\section{Challenge Results}
\label{sec:results}

Tables~\ref{table:1}--\ref{table:5} summarize the leader boards for Tracks 1--5, respectively.

\subsection{Summary for the Track 1 Challenge}

\begin{table}[t]
\caption{Summary of the Track 1 leader board.}
\label{table:1}
\centering
\footnotesize
\begin{tabular}{|c|c|c|c|c|}
\hline
Rank & Team ID & Team & Score (HOTA) & Online \\
\hline\hline
1 & 221 & Yachiyo~\cite{AICity24Paper32} & {\bf 71.9446} & No \\
\hline
2 & 79 & SJTU-Lenovo~\cite{AICity24Paper1} & 67.2175 & Yes \\
\hline
3 & 40 & Nota~\cite{AICity24Paper39} & 60.9261 & Yes \\
\hline
4 & 142 & Fraunhofer IOSB~\cite{AICity24Paper46} & 60.8792 & Yes \\
\hline
5 & 8 & UW-ETRI~\cite{AICity24Paper6} & 57.1445 & Yes \\
\hline
6 & 50 & ARV~\cite{AICity24Paper40} & 51.0556 & Yes \\
\hline
9 & 162 & Asilla~\cite{AICity24Paper43} & 40.3361 & Yes \\
\hline
\end{tabular}
\vspace{-0.4cm}
\end{table}

The teams all employed state-of-the-art YOLO-based models for person detection, notably YOLOX~\cite{yolox2021} and YOLOv8~\cite{yolov8_ultralytics}. For re-identification (ReID), they continued to use advanced models similar to those in previous years, including OSNet~\cite{zhou2020omni}, TransReID~\cite{wu2021self}, and their combinations. Fraunhofer IOSB~\cite{AICity24Paper46} utilized a transformer-based model~\cite{chen2023beyond} that was pre-trained on large-scale data~\cite{zhong2019unsupervised}. Since the evaluation required 3D locations, almost all teams implemented pose estimation to accurately determine foot positions. The top three teams adopted HRNet~\cite{sun2019deep}. The UW-ETRI team~\cite{AICity24Paper6} trained a YOLO-based model for joint and keypoint detection that was more computation-efficient.

Regarding single-camera tracking, most teams leveraged established state-of-the-art methods, such as BoT-SORT~\cite{jiang2020bot}, StrongSORT~\cite{Du_2021_ICCV}, ByteTrack~\cite{wang2021box}, and ConfTrack~\cite{jung2024conftrack}. The top team~\cite{AICity24Paper32} proposed an Overlap Suppression Clustering scheme to generate non-overlapping tracklets in single-camera setups.

This year, most teams were encouraged to adopt online methods for multi-camera tracking, which are more suitable for real-time applications. These methods maintain a global state of "anchors" derived from past tracking results, updating these anchors based on new data in the current time window. Various schemes were introduced to correct false positives, negatives, and ID switches during online tracking. For instance, the Nota team~\cite{AICity24Paper39} implemented Appearance Feature Refinement using agglomerative clustering to update the appearance features for each anchor, and Overlapped Cluster Refinement to solve duplicate assignments. The ARV team~\cite{AICity24Paper40} also applied hierarchical clustering with appearance features and spatio-temporal constraints, enhancing accuracy with spatio-temporal refinement and cross-interval synchronization.

Despite these advancements in online tracking, the offline method by Yachiyo~\cite{AICity24Paper32} still showed significant advantages. Their approach involved extracting representative images from each tracklet. ReID was performed only on images identified as highly recognizable through pose estimation. Tracklets composed solely of low-identifiable images were assigned to separate clusters in the ReID process.

\subsection{Summary for the Track 2 Challenge}

\begin{table}[t]
  \caption{Summary of the Track 2 leader board.}
  \label{table:2}
  \centering
  \footnotesize
  \begin{tabular}{|c|c|c|c|}
    \hline
    Rank & Team ID & Team & Score (4 metrics avg.) \\
    \hline\hline
    1 & 208 & AliOpenTrek~\cite{AICity24Paper38} & {\bf 33.4308} \\
    \hline
    2 & 28 & AIO\_ISC~\cite{AICity24Paper10} & 32.8877 \\
    \hline
    3 & 68 & Lighthouse~\cite{AICity24Paper28} & 32.3006 \\
    \hline
    6 & 219 & UCF-SST-NLP~\cite{AICity24Paper27} & 29.0084 \\
    \hline
    9 & 91 & HCMUS\_AGAIN~\cite{AICity24Paper15} & 22.7371 \\
    \hline
  \end{tabular}
\end{table}

Track 2 features a detailed video captioning task within traffic videos. Most teams~\cite{AICity24Paper38,AICity24Paper10,AICity24Paper28, AICity24Paper15} employed Vision Language Model (VLM) based methods, with Teams~\cite{AICity24Paper38,AICity24Paper10,AICity24Paper15} using Large Language Models (LLMs) as text decoders. These teams used LLMs to generate captions by processing inputs through vision and text encoders. Notable VLMs employed by the first- and second-place teams included LLaVA-1.6-34B~\cite{liu2024llavanext}, Qwen-VL~\cite{Qwen-VL}, and Video-LLaVA~\cite{lin2023videollava}, while the LLM components utilized were Vicuna~\cite{vicuna2023} and Qwen~\cite{qwen}.

Team~\cite{AICity24Paper28} utilized a Vid2Seq~\cite{yang2023vid2seq} based approach with a T5-Base~\cite{raffel2023exploring} text decoder. For vision encoding, all teams using VLM methodologies opted for CLIP ViT-L/14~\cite{radford2021learning}.

Further innovations were seen with Teams~\cite{AICity24Paper38,AICity24Paper28} proposing the simultaneous use of global and local views to enhance performance, with Team~\cite{AICity24Paper28} achieving significant improvements through temporal modeling of local features. Interestingly, Team~\cite{AICity24Paper38} explored using Reinforcement Learning from Human Feedback (RLHF), although this approach did not perform well due to the challenges in aligning with the diverse and lengthy description patterns.

A novel visual prompt schema aimed at domain-specific task utilization was proposed by Team~\cite{AICity24Paper38}, facilitating the creation of instruction data. Team~\cite{AICity24Paper10} focused on extracting hierarchical structures from captions as a preprocessing step, implementing a two-stage training process to enhance the accuracy of segment and description generation.

Multi-view information was leveraged by Team~\cite{AICity24Paper15} to improve results through several perception-based approaches within a rule engine framework. Additionally, Team~\cite{AICity24Paper27} explored knowledge transfer across different traffic domains, initially training with annotations from the BDD dataset before fine-tuning on the WTS staged dataset.

\subsection{Summary for the Track 3 Challenge}

\begin{table}[t]
\caption{Summary of the Track 3 leader board.}
\label{table:3}
\centering
\footnotesize
\begin{tabular}{|c|c|c|c|}
\hline
Rank & Team ID & Team & Score (activity overlap score) \\
\hline\hline
1 & 155& TeleAl~\cite{AICity24Paper24}& {\bf 0.8282} \\
\hline
5& 5& SKKU-AutoLab~\cite{AICity24Paper31} & 0.7798\\
\hline
8 & 165 & MCPRL~\cite{AICity24Paper37} & 0.6080 \\\hline
\end{tabular}
\vspace{-0.4cm}
\end{table}

The top-performing teams in Track 3 of the Challenge focused on methodologies centered around activity recognition, specifically addressing two key aspects: (1) classifying various distracted driving activities, and (2) Temporal Action Localization (TAL), which determines the start and end times for each activity. The leading team, TeleAl \cite{AICity24Paper24}, developed an Augmented Self-Mask Attention (AMA) architecture that enhanced the learning of bidirectional contexts, resulting in improved handling of overlapping TALs. They further enhanced their approach by applying an ensemble method combined with weighted boundaries fusion, which helped in identifying TALs with high confidence levels. Their best score was 0.8282.

The second-place team~\cite{AICity24Paper31} focused on large model fine-tuning and used ensemble methods to achieve clip-level classification for short video segments. To refine TAL, they employed a multi-step post-processing algorithm that enhanced the precision of activity boundaries.


Team~\cite{AICity24Paper37} built a multi-view fusion and adaptive thresholding algorithm to address the challenges posed by similar action behaviors and interference from background activity. For their TAL approach, they designed a post-processing procedure that enabled fine localization from initially coarse estimates through techniques such as post-connection and candidate behavior merging.

Lastly, Team~\cite{AICity24Paper51} leveraged Graph-Based Change-Point Detection to generate action proposals, alongside a Video Large Language Model (Video-LLM) for robust activity recognition.

\subsection{Summary for the Track 4 Challenge}

\begin{table}[t]
\caption{Summary of the Track 4 leader board.}
\label{table:4}
\centering
\footnotesize
\begin{tabular}{|c|c|c|c|}
\hline
Rank & Team ID & Team         & Score (F1) \\ \hline
1    & 9       & VNPT AI~\cite{AICity24Paper4}    & 0.6406     \\ \hline
2    & 40      & Nota~\cite{AICity24Paper50}   & 0.6196     \\ \hline
3    & 5       & SKKU-AutoLab~\cite{AICity24Paper20} & 0.6194     \\ \hline
4    & 63      & UIT\_AICLUB~\cite{AICity24Paper44}  & 0.6077     \\ \hline
5    & 15      & SKKU-NDSU~\cite{AICity24Paper12}   & 0.5965     \\ \hline
6    & 33      & MCPRL~\cite{AICity24Paper19}       & 0.5883     \\ \hline
\end{tabular}
\vspace{-0.4cm}
\end{table}

Most teams~\cite{AICity24Paper4, AICity24Paper50, AICity24Paper44, AICity24Paper12, AICity24Paper19} employed an ensemble model~\cite{WBF} to enhance their model performance and generalization capabilities. The winning team, VNPT AI~\cite{AICity24Paper4}, integrated multiple models including CO-DETR~\cite{Co-DETR}, YOLOv9~\cite{YOLOv9}, YOLOR-W6~\cite{YOLOR-W6}, and InternImage~\cite{InternImage}, alongside pseudo labels generated from pre-trained models on various combinations of the FishEye8K~\cite{Gochoo_2023_CVPR} and VisDrone~\cite{Visdrone} datasets. Their approach achieved the highest F1 score of 0.6604 among all participants in Track 4.

The runner-up, Nota~\cite{AICity24Paper50}, employed DINO~\cite{Dino} with ViT-L~\cite{ViT-L} and Swin-L~\cite{swin-L} backbones, supplemented with other techniques such as StableSR~\cite{SR} and histogram equalization. The technique of Slicing Aided Hyper Inference (SAHI)~\cite{SAHI} was utilized by both Nota and UIT\textunderscore AICLUB~\cite{AICity24Paper44}, the fourth-place team, to enhance detection of distorted and blurred small objects, a common challenge with fisheye lenses.

The third-place team, SKKU-AutoLab~\cite{AICity24Paper20}, developed a synthetic dataset using CycleGAN~\cite{Cycle-GAN} and pseudo labels generated by the YOLO-World~\cite{YOLO-World} model, training their YOLOR-D6~\cite{YOLOR} model on this dataset to achieve a score of 0.6194.

Additionally, the fifth-place team, SKKU-NDSU~\cite{AICity24Paper12}, proposed a Low-Light Image Enhancement Framework that converts night-time images into daylight-like images using GSAD~\cite{GSAD}, creating a unified dataset. For post-processing, they employed super-resolution techniques using DAT~\cite{DAT} during the testing phase.

Lastly, the team MCPRL~\cite{AICity24Paper19} introduced post-processing modules named static object processing and confidence score refinement. This method differentiates static objects across sequential frames, refining detection by excluding static false positives and incorporating overlooked false negatives.

\subsection{Summary for the Track 5 Challenge}

\begin{table}[t]
\caption{Summary of the Track 5 leader board.}
\label{table:5}
\centering
\footnotesize
\begin{tabular}{|c|c|c|c|}
\hline
Rank & Team ID & Team & Score (mAP) \\
\hline\hline
1 & 99 & UIT~\cite{AICity24Paper36} & {\bf 0.4860} \\
\hline
2 & 76 & China Mobile~\cite{AICity24Paper17} & 0.4824 \\
\hline
3 & 9 & VNPT~\cite{AICity24Paper5} & 0.4792 \\
\hline
7 & 57 & BUPT~\cite{AICity24Paper13} & 0.394 \\
\hline
\end{tabular}
\end{table}

In Track 5, focusing on object detection, most teams employed a combination of object detection and multiple object tracking techniques. These approaches typically involve several key components:

\textbf{Object Detection:} Most teams utilized state-of-the-art Transformer models combined with ensemble techniques. The top-performing team~\cite{AICity24Paper36} primarily used Co-DETR~\cite{zong2023detrs} for object detection, while the second-ranked team~\cite{AICity24Paper17} implemented Co-DETR in conjunction with the DETA algorithm~\cite{deta} to refine bounding box localization. The third-ranked team~\cite{AICity24Paper5} deployed separate sub-modules for vehicle and person detection and an additional one for head detection. For vehicle and person detection, they used YoloV7~\cite{YoloV7}, YoloV8~\cite{yolov8_ultralytics}, and Co-DETR with a Swin-L backbone~\cite{zong2023detrs}. For head detection, they integrated a Swin-L backbone into the Co-DETR architecture.

\textbf{Ensemble Techniques:} The top team employed Weighted Box Fusion~\cite{solovyev2021weighted}, while the second-ranked team combined weighted box fusion (WBF) with non-maximum suppression (NMS)~\cite{NMS} and Test Time Augmentation (TTA)~\cite{TTA}. Similarly, the third team also used WBF and TTA to enhance detection accuracy.

\textbf{Handling Class Imbalance:} A major challenge across the teams was dealing with class imbalance within the dataset. The first-ranked team addressed this by employing the Minority Optimizer Algorithm~\cite{AICity24Paper36} to improve recall for rare classes, prioritizing thresholds for these classes to maintain robust recall. They also developed a strategy to balance precision and recall, creating virtual bounding boxes with calibrated confidence scores to optimize recall where the detector failed to classify objects accurately. The third-ranked team used an object association module~\cite{AICity24Paper5} for pairing humans with motorbikes and heads and applied a tracking module to ascertain vehicle direction. They implemented a confidence score correction scheme to adjust for class imbalances. The second-ranked team augmented the dataset with general image processing techniques, randomly cropping and resizing augmented inputs to enable multi-scale object detection.

%% file: 6Conclusion.tex
\section{Discussion and Conclusion}
\label{sec:conclusion}

The 8th AI City Challenge has continued to attract substantial interest from the global research community, evidenced by both the quantity and the quality of the participants. We wish to highlight several notable insights from the event.

In Track 1, we enhanced the benchmark for MTMC people tracking by expanding the scale and improving evaluation metrics, emphasizing online methods this year. While an offline method has attained nearly a 72\% HOTA on this extensive dataset, the top-performing online method~\cite{AICity24Paper1} only achieved approximately 67\%. Before these methods can be effectively utilized in real-world applications, there are several challenges to overcome. First, most teams deployed separate models for detection and pose estimation, some of which are based on computationally intensive transformer models. Second, despite numerous proposed schemes to refine trajectories in multi-camera tracking, these methods predominantly remain rule-based and do not exploit the large-scale MTMC data. We encourage teams to investigate learning-based tracking methods using Graph Neural Networks (GNNs) or other pertinent architectures. Third, certain teams presupposed a known number of individuals, an assumption that may not hold in practical settings. They must develop strategies to manage the dynamics of individuals entering and exiting the scene. For future challenges, we plan to incorporate datasets featuring individuals in similar attire, which, although challenging, reflects common scenarios in warehouses and sporting events.

Track 2 presents unique challenges, primarily how methods adapt to the traffic domain video data, which significantly differs from more common public datasets. Another challenge is accurately generating detailed and lengthy descriptions from video at the instance level. Participants widely utilized large VLMs for deep video-language understanding. Despite their strong generalization capabilities, LLMs face challenges in domain-specific data, particularly in detailing traffic scenarios linguistically. Traditional metrics such as BLEU, METEOR, ROUGE, and CIDEr focus on syntactic similarity but struggle to assess semantic accuracy in lengthy, detailed captions. We encourage teams to explore VLM designs focusing on spatial-temporal relationships at the instance level to enhance task performance.

In Track 3, teams engaged with the expanded SynDD2 benchmark~\cite{rahman2023synthetic} to tackle the Driver Activity Recognition challenge. This involved classifying driver activities and localizing them temporally to determine their start and end times. Efforts included developing specialized architectures, optimizing algorithms, and crafting pipelines to boost detection efficiency. Techniques employed included prompt engineering with language models~\cite{yang2019xlnet, maaz202XvideochatGPT}, vision transformers~\cite{10.1007/978-3-031-19772-7_29, liu2022video}, and action classifiers~\cite{feichtenhofer2020x3d, tong2022videomae, wang2023videomae, li2023uniformer, li2022uniformerv2}. Ongoing challenges in activity recognition and temporal action localization highlight the need for further research and more refined datasets.

The majority of teams in Track 4 utilized ensemble models to enhance performance and generalization. The winning team implemented a combination of CO-DETR, YOLOv9, YOLOR-W6, and InternImage models, supplemented by pseudo labels from the FishEye8K and VisDrone datasets, achieving an F1 score of 0.6604. Other notable approaches included employing DINO with ViT-L and Swin-L backbones, StableSR, and histogram equalization. Techniques such as SAHI addressed challenges related to fisheye lens distortion. Innovations like synthetic image generation using CycleGAN and enhancing images under low-light conditions with specialized frameworks and post-processing techniques underscored the diverse strategies teams used to tackle complex vision tasks.

In Track 5, teams were provided with a challenging dataset for motorbike helmet violation detection in an Indian city. The state-of-the-art model achieved a 0.4860 mAP~\cite{AICity24Paper36}. Top teams employed advanced object detection models such as Co-DETR~\cite{zong2023detrs} alongside ensembling techniques and class enhancement strategies to improve accuracy and model performance.

%% file: 7Acknowledgement.tex
\section{Acknowledgment}

The datasets for the 8th AI City Challenge were developed through extensive data curation efforts. This was made possible by the contributions from both industry and academia. Notable contributors include NVIDIA Corporation and Woven by Toyota, Inc., along with significant academic collaborations. These academic partners comprised Iowa State University, National Yang Ming Chiao Tung University, Indian Institute of Technology Kanpur, United Arab Emirates University, and the Emirates Center for Mobility Research (ECMR), which supported the initiative through Grant 12R012.

%% file: main.bbl
\begin{thebibliography}{10}\itemsep=-1pt

\bibitem{SAHI}
Fatih~Cagatay Akyon, Sinan Onur~Altinuc, and Alptekin Temizel.
\newblock Slicing aided hyper inference and fine-tuning for small object detection.
\newblock In {\em 2022 IEEE International Conference on Image Processing (ICIP)}. IEEE, Oct. 2022.

\bibitem{qwen}
Jinze Bai, Shuai Bai, Yunfei Chu, Zeyu Cui, Kai Dang, Xiaodong Deng, Yang Fan, Wenbin Ge, Yu Han, Fei Huang, Binyuan Hui, Luo Ji, Mei Li, Junyang Lin, Runji Lin, Dayiheng Liu, Gao Liu, Chengqiang Lu, Keming Lu, Jianxin Ma, Rui Men, Xingzhang Ren, Xuancheng Ren, Chuanqi Tan, Sinan Tan, Jianhong Tu, Peng Wang, Shijie Wang, Wei Wang, Shengguang Wu, Benfeng Xu, Jin Xu, An Yang, Hao Yang, Jian Yang, Shusheng Yang, Yang Yao, Bowen Yu, Hongyi Yuan, Zheng Yuan, Jianwei Zhang, Xingxuan Zhang, Yichang Zhang, Zhenru Zhang, Chang Zhou, Jingren Zhou, Xiaohuan Zhou, and Tianhang Zhu.
\newblock Qwen technical report.
\newblock {\em arXiv preprint arXiv:2309.16609}, 2023.

\bibitem{Qwen-VL}
Jinze Bai, Shuai Bai, Shusheng Yang, Shijie Wang, Sinan Tan, Peng Wang, Junyang Lin, Chang Zhou, and Jingren Zhou.
\newblock Qwen-vl: A versatile vision-language model for understanding, localization, text reading, and beyond.
\newblock {\em arXiv preprint arXiv:2308.12966}, 2023.

\bibitem{banerjee-lavie-2005-meteor}
Satanjeev Banerjee and Alon Lavie.
\newblock {METEOR}: An automatic metric for {MT} evaluation with improved correlation with human judgments.
\newblock In Jade Goldstein, Alon Lavie, Chin-Yew Lin, and Clare Voss, editors, {\em Proceedings of the {ACL} Workshop on Intrinsic and Extrinsic Evaluation Measures for Machine Translation and/or Summarization}, pages 65--72, Ann Arbor, Michigan, June 2005. Association for Computational Linguistics.

\bibitem{chen2023beyond}
Weihua Chen, Xianzhe Xu, Jian Jia, Hao Luo, Yaohua Wang, Fan Wang, Rong Jin, and Xiuyu Sun.
\newblock Beyond appearance: a semantic controllable self-supervised learning framework for human-centric visual tasks.
\newblock In {\em The IEEE/CVF Conference on Computer Vision and Pattern Recognition}, 2023.

\bibitem{AICity24Paper17}
Yunliang Chen, Chen Wang, and Yingda Shang.
\newblock An effective method for detecting violation of helmet rule for motorcyclists.
\newblock In {\em CVPR Workshop}, Seattle, WA, USA, 2024.

\bibitem{DAT}
Zheng Chen, Yulun Zhang, Jinjin Gu, Linghe Kong, Xiaokang Yang, and Fisher Yu.
\newblock Dual aggregation transformer for image super-resolution, 2023.

\bibitem{YOLO-World}
Tianheng Cheng, Lin Song, Yixiao Ge, Wenyu Liu, Xinggang Wang, and Ying Shan.
\newblock Yolo-world: Real-time open-vocabulary object detection.
\newblock In {\em Proc. IEEE Conf. Computer Vision and Pattern Recognition (CVPR)}, 2024.

\bibitem{vicuna2023}
Wei-Lin Chiang, Zhuohan Li, Zi Lin, Ying Sheng, Zhanghao Wu, Hao Zhang, Lianmin Zheng, Siyuan Zhuang, Yonghao Zhuang, Joseph~E. Gonzalez, Ion Stoica, and Eric~P. Xing.
\newblock Vicuna: An open-source chatbot impressing gpt-4 with 90\%* chatgpt quality, March 2023.

\bibitem{AICity24Paper28}
Quang~Minh Dinh, Minh~Khoi Ho, Quan~Anh Dang, and Phong Ngoc~Hung Tran.
\newblock Trafficvlm: A controllable visual language model for traffic video captioning.
\newblock In {\em CVPR Workshop}, Seattle, WA, USA, 2024.

\bibitem{ViT-L}
Alexey Dosovitskiy, Lucas Beyer, Alexander Kolesnikov, Dirk Weissenborn, Xiaohua Zhai, Thomas Unterthiner, Mostafa Dehghani, Matthias Minderer, Georg Heigold, Sylvain Gelly, Jakob Uszkoreit, and Neil Houlsby.
\newblock An image is worth 16x16 words: Transformers for image recognition at scale, 2021.

\bibitem{Du_2021_ICCV}
Yunhao Du, Junfeng Wan, Yanyun Zhao, Binyu Zhang, Zhihang Tong, and Junhao Dong.
\newblock Giaotracker: A comprehensive framework for mcmot with global information and optimizing strategies in visdrone 2021.
\newblock In {\em Proceedings of the IEEE/CVF International Conference on Computer Vision (ICCV) Workshops}, pages 2809--2819, October 2021.

\bibitem{AICity24Paper38}
Zhizhao Duan, Hao Cheng, Xu Duo, Xi Wu, Xiangxie Zhang, Ye Xi, and Zhen Xie.
\newblock Cityllava: Efficient fine-tuning for vlms in city scenario.
\newblock In {\em CVPR Workshop}, Seattle, WA, USA, 2024.

\bibitem{AICity24Paper4}
Hung~Viet Duong, Quyen~Duc Nguyen, Thien~Van Luong, Huan Vu, and Cuong~Tien Nguyen.
\newblock Robust data augmentation and ensemble method for object detection in fisheye camera images.
\newblock In {\em CVPR Workshop}, Seattle, WA, USA, 2024.

\bibitem{pascal-voc-2012}
M. Everingham, L. Van~Gool, C.~K.~I. Williams, J. Winn, and A. Zisserman.
\newblock The {PASCAL} {V}isual {O}bject {C}lasses {C}hallenge 2012 {(VOC2012)} {R}esults.
\newblock http://www.pascal-network.org/challenges/VOC/voc2012/workshop/index.html.

\bibitem{feichtenhofer2020x3d}
Christoph Feichtenhofer, Haoqi Fan, Jitendra Malik, and Kaiming He.
\newblock X3d: Expanding architectures for efficient video recognition.
\newblock {\em arXiv preprint arXiv:2004.04730}, 2020.

\bibitem{yolox2021}
Zheng Ge, Songtao Liu, Feng Wang, Zeming Li, and Jian Sun.
\newblock Yolox: Exceeding yolo series in 2021.
\newblock {\em arXiv preprint arXiv:2107.08430}, 2021.

\bibitem{geiger2012we}
Andreas Geiger, Philip Lenz, and Raquel Urtasun.
\newblock Are we ready for autonomous driving? the kitti vision benchmark suite.
\newblock In {\em 2012 IEEE conference on computer vision and pattern recognition}, pages 3354--3361. IEEE, 2012.

\bibitem{AICity24Paper44}
Bao~Tran Gia, Tuong Bui~Cong Khanh, Hien~Trong Ho, Thuyen~Tran Doan, Tien Do, Duy-Dinh Le, and Thanh~Duc Ngo.
\newblock Enhancing road object detection in fisheye cameras: An effective framework integrating sahi and hybrid inference.
\newblock In {\em CVPR Workshop}, Seattle, WA, USA, 2024.

\bibitem{NMS}
Ross Girshick.
\newblock Fast r-cnn.
\newblock In {\em Proceedings of the IEEE international conference on computer vision}, pages 1440--1448, 2015.

\bibitem{Gochoo_2023_CVPR}
Munkhjargal Gochoo, Munkh-Erdene Otgonbold, Erkhembayar Ganbold, Jun-Wei Hsieh, Ming-Ching Chang, Ping-Yang Chen, Byambaa Dorj, Hamad Al~Jassmi, Ganzorig Batnasan, Fady Alnajjar, Mohammed Abduljabbar, and Fang-Pang Lin.
\newblock Fisheye8k: A benchmark and dataset for fisheye camera object detection.
\newblock In {\em Proceedings of the IEEE/CVF Conference on Computer Vision and Pattern Recognition (CVPR) Workshops}, pages 5304--5312, June 2023.

\bibitem{UAEU23FishEye8K}
Munkhjargal Gochoo, Munkh-Erdene Otgonbold, Erkhembayar Ganbold, Jun-Wei Hsieh, Ming-Ching Chang, Ping-Yang Chen, Byambaa Dorj, Hamad~Al Jassmi, Ganzorig Batnasan, Fady Alnajjar, Mohammed Abduljabbar, and Fang-Pang Lin.
\newblock {FishEye8K}: {A} benchmark and dataset for fisheye camera object detection.
\newblock In {\em CVPR Workshop}, 2023.

\bibitem{GSAD}
Jinhui Hou, Zhiyu Zhu, Junhui Hou, Hui Liu, Huanqiang Zeng, and Hui Yuan.
\newblock Global structure-aware diffusion process for low-light image enhancement, 2023.

\bibitem{jiang2020bot}
Zhuangzhi Jiang, Zhipeng Ye, Junzhou Huang, Jian Zheng, and Jian Zhang.
\newblock Bot-sort: Robust associations multi-pedestrian tracking.
\newblock In {\em Proceedings of the IEEE/CVF Conference on Computer Vision and Pattern Recognition Workshops}, pages 702--703, 2020.

\bibitem{yolov8_ultralytics}
Glenn Jocher, Ayush Chaurasia, and Jing Qiu.
\newblock Ultralytics yolov8, 2023.

\bibitem{luiten2020trackeval}
Arne~Hoffhues Jonathon~Luiten.
\newblock Trackeval.
\newblock \url{https://github.com/JonathonLuiten/TrackEval}, 2020.

\bibitem{jung2024conftrack}
Hyeonchul Jung, Seokjun Kang, Takgen Kim, and HyeongKi Kim.
\newblock Conftrack: Kalman filter-based multi-person tracking by utilizing confidence score of detection box.
\newblock In {\em Proceedings of the IEEE/CVF Winter Conference on Applications of Computer Vision}, pages 6583--6592, 2024.

\bibitem{AICity24Paper39}
Jeongho Kim, Wooksu Shin, Hancheol Park, and Donghyuk Choi.
\newblock Cluster self-refinement for enhanced online multi-camera people tracking.
\newblock In {\em CVPR Workshop}, Seattle, WA, USA, 2024.

\bibitem{WTS2024}
Quan Kong, Yuki Kawana, Rajat Saini, Ashutosh Kumar, Jingjing Pan, Ta Gu, Yohei Ozao, Balazs Opra, David~C. Anastasiu, Yoichi Sato, and Norimasa Kobori.
\newblock Wts: A pedestrian-centric traffic video dataset for fine-grained spatial-temporal understanding.
\newblock 2024.

\bibitem{li2022uniformerv2}
Kunchang Li, Yali Wang, Yinan He, Yizhuo Li, Yi Wang, Limin Wang, and Yu Qiao.
\newblock Uniformerv2: Spatiotemporal learning by arming image vits with video uniformer.
\newblock {\em arXiv preprint arXiv:2211.09552}, 2022.

\bibitem{li2023uniformer}
Kunchang Li, Yali Wang, Junhao Zhang, Peng Gao, Guanglu Song, Yu Liu, Hongsheng Li, and Yu Qiao.
\newblock Uniformer: Unifying convolution and self-attention for visual recognition.
\newblock {\em IEEE Transactions on Pattern Analysis and Machine Intelligence}, 2023.

\bibitem{lin2023videollava}
Bin Lin, Yang Ye, Bin Zhu, Jiaxi Cui, Munan Ning, Peng Jin, and Li Yuan.
\newblock Video-llava: Learning united visual representation by alignment before projection, 2023.

\bibitem{lin-2004-rouge}
Chin-Yew Lin.
\newblock {ROUGE}: A package for automatic evaluation of summaries.
\newblock In {\em Text Summarization Branches Out}, pages 74--81, Barcelona, Spain, July 2004. Association for Computational Linguistics.

\bibitem{liu2024llavanext}
Haotian Liu, Chunyuan Li, Yuheng Li, Bo Li, Yuanhan Zhang, Sheng Shen, and Yong~Jae Lee.
\newblock Llava-next: Improved reasoning, ocr, and world knowledge, January 2024.

\bibitem{swin-L}
Ze Liu, Yutong Lin, Yue Cao, Han Hu, Yixuan Wei, Zheng Zhang, Stephen Lin, and Baining Guo.
\newblock Swin transformer: Hierarchical vision transformer using shifted windows.
\newblock In {\em 2021 IEEE/CVF International Conference on Computer Vision (ICCV)}. IEEE, Oct. 2021.

\bibitem{liu2022video}
Ze Liu, Jia Ning, Yue Cao, Yixuan Wei, Zheng Zhang, Stephen Lin, and Han Hu.
\newblock Video swin transformer.
\newblock In {\em Proceedings of the IEEE/CVF conference on computer vision and pattern recognition}, pages 3202--3211, 2022.

\bibitem{luiten2020IJCV}
Jonathon Luiten, Aljosa Osep, Patrick Dendorfer, Philip Torr, Andreas Geiger, Laura Leal-Taix{\'e}, and Bastian Leibe.
\newblock Hota: A higher order metric for evaluating multi-object tracking.
\newblock {\em International Journal of Computer Vision}, pages 1--31, 2020.

\bibitem{AICity24Paper19}
Xingshuang Luo, Zhe Cui, and Fei Su.
\newblock Fe-det: An effective traffic object detection framework for fish-eye cameras.
\newblock In {\em CVPR Workshop}, Seattle, WA, USA, 2024.

\bibitem{AICity24Paper5}
Thien~Van Luong, Phúc Sĩ~Nguyễn Hữu, Khanh~Duy Dinh, Hung~Viet Duong, Sam Duy~Hong Vo, Huan Vu, Hoang~Minh Tuan, and Cuong~Tien Nguyen.
\newblock Motorcyclist helmet violation detection framework by leveraging robust ensemble and augmentation methods.
\newblock In {\em CVPR Workshop}, Seattle, WA, USA, 2024.

\bibitem{maaz202XvideochatGPT}
Muhammad Maaz, Hanoona Rasheed, Salman Khan, and Fahad~Shahbaz Khan.
\newblock Video-chatgpt: Towards detailed video understanding via large vision and language models.
\newblock {\em arXiv preprint arXiv:XXXX.XXXXX}, 202X.

\bibitem{Naphade18AIC18}
Milind Naphade, Ming-Ching Chang, Anuj Sharma, David~C. Anastasiu, Vamsi Jagarlamudi, Pranamesh Chakraborty, Tingting Huang, Shuo Wang, Ming-Yu Liu, Rama Chellappa, Jenq-Neng Hwang, and Siwei Lyu.
\newblock The 2018 {NVIDIA} {AI} {C}ity {C}hallenge.
\newblock In {\em CVPR Workshop}, pages 53--–60, 2018.

\bibitem{Naphade19AIC19}
Milind Naphade, Zheng Tang, Ming-Ching Chang, David~C. Anastasiu, Anuj Sharma, Rama Chellappa, Shuo Wang, Pranamesh Chakraborty, Tingting Huang, Jenq-Neng Hwang, and Siwei Lyu.
\newblock The 2019 {AI} {C}ity {C}hallenge.
\newblock In {\em CVPR Workshop}, page 452–460, 2019.

\bibitem{9857426}
M. Naphade, S. Wang, D.~C. Anastasiu, Z. Tang, M. Chang, Y. Yao, L. Zheng, M.~Shaiqur Rahman, A. Venkatachalapathy, A. Sharma, Q. Feng, V. Ablavsky, S. Sclaroff, P. Chakraborty, A. Li, S. Li, and R. Chellappa.
\newblock The 6th ai city challenge.
\newblock In {\em 2022 IEEE/CVF Conference on Computer Vision and Pattern Recognition Workshops (CVPRW)}, pages 3346--3355, Los Alamitos, CA, USA, jun 2022. IEEE Computer Society.

\bibitem{Naphade21AIC21}
Milind Naphade, Shuo Wang, David~C. Anastasiu, Zheng Tang, Ming-Ching Chang, Xiaodong Yang, Yue Yao, Liang Zheng, Pranamesh Chakraborty, Christian~E. Lopez, Anuj Sharma, Qi Feng, Vitaly Ablavsky, and Stan Sclaroff.
\newblock The 5th {AI City Challenge}.
\newblock In {\em The IEEE Conference on Computer Vision and Pattern Recognition (CVPR) Workshops}, June 2021.

\bibitem{Naphade20AIC20}
Milind Naphade, Shuo Wang, David~C. Anastasiu, Zheng Tang, Ming-Ching Chang, Xiaodong Yang, Liang Zheng, Anuj Sharma, Rama Chellappa, and Pranamesh Chakraborty.
\newblock The 4th {AI} {C}ity {C}hallenge.
\newblock In {\em CVPR Workshop}, 2020.

\bibitem{Naphade23AIC23}
Milind Naphade, Shuo Wang, David~C. Anastasiu, Zheng Tang, Ming-Ching Chang, Yue Yao, Liang Zheng, Mohammed~Shaiqur Rahman, Meenakshi~S. Arya, Anuj Sharma, Qi Feng, Vitaly Ablavsky, Stan Sclaroff, Pranamesh Chakraborty, Sanjita Prajapati, Alice Li, Shangru Li, Krishna Kunadharaju, Shenxin Jiang, and Rama Chellappa.
\newblock The 7th {AI City Challenge}.
\newblock In {\em The IEEE Conference on Computer Vision and Pattern Recognition (CVPR) Workshops}, June 2023.

\bibitem{Naphade22AIC22}
Milind Naphade, Shuo Wang, David~C. Anastasiu, Zheng Tang, Ming-Ching Chang, Yue Yao, Liang Zheng, Mohammed~Shaiqur Rahman, Archana Venkatachalapathy, Anuj Sharma, Qi Feng, Vitaly Ablavsky, Stan Sclaroff, Pranamesh Chakraborty, Alice Li, Shangru Li, and Rama Chellappa.
\newblock The 6th {AI City Challenge}.
\newblock In {\em The IEEE Conference on Computer Vision and Pattern Recognition (CVPR) Workshops}, June 2022.

\bibitem{AICity24Paper31}
Huy-Hung Nguyen, TRAN~DAI CHI, Long~Hoang Pham, Duong Nguyen-Ngoc Tran, Tai Huu~Phuong Tran, Duong~Khac Vu, Quoc Pham~Nam Ho, Ngoc Doan-Minh Huynh, Hyung-Min Jeon, Hyung-Joon Jeon, and Jae Jeon.
\newblock Multi-view spatial-temporal learning for understanding unusual behaviors in untrimmed naturalistic driving videos.
\newblock In {\em CVPR Workshop}, Seattle, WA, USA, 2024.

\bibitem{oliveira-dos-santos-etal-2021-cider}
Gabriel Oliveira~dos Santos, Esther~Luna Colombini, and Sandra Avila.
\newblock {CIDE}r-{R}: Robust consensus-based image description evaluation.
\newblock In Wei Xu, Alan Ritter, Tim Baldwin, and Afshin Rahimi, editors, {\em Proceedings of the Seventh Workshop on Noisy User-generated Text (W-NUT 2021)}, pages 351--360, Online, Nov. 2021. Association for Computational Linguistics.

\bibitem{gpt3.5}
OpenAI.
\newblock {GPT-3.5}, 2023.

\bibitem{deta}
Jeffrey Ouyang-Zhang, Jang~Hyun Cho, Xingyi Zhou, and Philipp Kr{\"a}henb{\"u}hl.
\newblock Nms strikes back.
\newblock {\em arXiv preprint arXiv:2212.06137}, 2022.

\bibitem{papineni-etal-2002-bleu}
Kishore Papineni, Salim Roukos, Todd Ward, and Wei-Jing Zhu.
\newblock {B}leu: a method for automatic evaluation of machine translation.
\newblock In Pierre Isabelle, Eugene Charniak, and Dekang Lin, editors, {\em Proceedings of the 40th Annual Meeting of the Association for Computational Linguistics}, pages 311--318, Philadelphia, Pennsylvania, USA, July 2002. Association for Computational Linguistics.

\bibitem{Visdrone}
{Pengfei Zhu}, {Dawei Du}, {Longyin Wen}, Xiao Bian, {Haibin Ling}, {Qinghua Hu}, Tao Peng, {Jiayu Zheng}, {Xinyao Wang}, Yue Zhang, {Liefeng Bo}, {Hailin Shi}, Rui Zhu, Bing Dong, {Dheeraj Reddy Pailla}, Feng Ni, {Guangyu Gao}, {Guizhong Liu}, {Haitao Xiong}, Jing Ge, {Jingkai Zhou}, {Jinrong Hu}, Lin Sun, Long Chen, Martin Lauer, Qiong Liu, Sai~Saketh Chennamsetty, Ting Sun, Tong Wu, Alex Kollerathu, Wei Tian, {Weida Qin}, {Xier Chen}, {Xingjie Zhao}, {Yanchao Lian}, {Yinan Wu}, Ying Li, {Yingping Li}, {Yiwen Wang}, {Yuduo Song}, {Yuehan Yao}, {Yunfeng Zhang}, {Zhaoliang Pi}, {Zhaotang Chen}, {Zhenyu Xu}, {Zhibin Xiao}, {Zhipeng Luo}, and {Ziming Liu}.
\newblock Visdrone-vid2019: The vision meets drone object detection in video challenge results.
\newblock 2019.

\bibitem{AICity24Paper20}
Long~Hoang Pham, Quoc Pham~Nam Ho, Duong Nguyen-Ngoc Tran, Tai Huu~Phuong Tran, Huy-Hung Nguyen, Duong~Khac Vu, TRAN~DAI CHI, Ngoc Doan-Minh Huynh, Hyung-Min Jeon, Hyung-Joon Jeon, and Jae Jeon.
\newblock Improving object detection to fisheye cameras with open-vocabulary pseudo-label approach.
\newblock In {\em CVPR Workshop}, Seattle, WA, USA, 2024.

\bibitem{radford2021learning}
Alec Radford, Jong~Wook Kim, Chris Hallacy, Aditya Ramesh, Gabriel Goh, Sandhini Agarwal, Girish Sastry, Amanda Askell, Pamela Mishkin, Jack Clark, Gretchen Krueger, and Ilya Sutskever.
\newblock Learning transferable visual models from natural language supervision, 2021.

\bibitem{raffel2023exploring}
Colin Raffel, Noam Shazeer, Adam Roberts, Katherine Lee, Sharan Narang, Michael Matena, Yanqi Zhou, Wei Li, and Peter~J. Liu.
\newblock Exploring the limits of transfer learning with a unified text-to-text transformer, 2023.

\bibitem{AICity24Paper51}
Mohammed~S Rahman, Anuj Sharma, Lynna Chu, and Ibne~Farabi Shihab.
\newblock Deeplocalization: Using change point detection for temporal action localization.
\newblock In {\em CVPR Workshop}, Seattle, WA, USA, 2024.

\bibitem{rahman2023synthetic}
Mohammed~Shaiqur Rahman, Jiyang Wang, Senem~Velipasalar Gursoy, David Anastasiu, Shuo Wang, and Anuj Sharma.
\newblock Synthetic distracted driving (syndd2) dataset for analyzing distracted behaviors and various gaze zones of a driver, 2023.

\bibitem{TTA}
Divya Shanmugam, Davis Blalock, Guha Balakrishnan, and John Guttag.
\newblock Better aggregation in test-time augmentation.
\newblock In {\em Proceedings of the IEEE/CVF international conference on computer vision}, pages 1214--1223, 2021.

\bibitem{AICity24Paper50}
Wooksu Shin, Donghyuk Choi, Hancheol Park, and Jeongho Kim.
\newblock Road object detection robust to distorted objects at the edge regions of images.
\newblock In {\em CVPR Workshop}, Seattle, WA, USA, 2024.

\bibitem{AICity24Paper27}
Maged Shoman, Dongdong Wang, Armstrong Aboah, and Mohamed Abdel-Aty.
\newblock Enhancing traffic safety with parallel dense video captioning for end-to-end event analysis.
\newblock In {\em CVPR Workshop}, Seattle, WA, USA, 2024.

\bibitem{WBF}
Roman Solovyev, Weimin Wang, and Tatiana Gabruseva.
\newblock Weighted boxes fusion: Ensembling boxes from different object detection models.
\newblock {\em Image and Vision Computing}, 107:104117, Mar. 2021.

\bibitem{solovyev2021weighted}
Roman Solovyev, Weimin Wang, and Tatiana Gabruseva.
\newblock Weighted boxes fusion: Ensembling boxes from different object detection models.
\newblock {\em Image and Vision Computing}, 107:104117, 2021.

\bibitem{AICity24Paper46}
Andreas Specker.
\newblock Ocmctrack: Online multi-target multi-camera tracking with corrective matching cascade.
\newblock In {\em CVPR Workshop}, Seattle, WA, USA, 2024.

\bibitem{sun2019deep}
Ke Sun, Bin Xiao, Dong Liu, and Jingdong Wang.
\newblock Deep high-resolution representation learning for human pose estimation.
\newblock In {\em CVPR}, 2019.

\bibitem{AICity24Paper40}
Vasin Suttichaya, Riu Cherdchusakulchai, Sasin Phimsiri, Visarut Trairattanapa, Suchat Tungjitnob, Wasu Kudisthalert, Pornprom Kiawjak, Ek Thamwiwatthana, Phawat Borisuitsawat, Teepakorn Tosawadi, Pakcheera Choppradit, Kasisdis Mahakijdechachai, Supawit Vatathanavaro, and Worawit Saetan.
\newblock Online multi-camera people tracking with spatial-temporal mechanism and anchor-feature hierarchical clustering.
\newblock In {\em CVPR Workshop}, Seattle, WA, USA, 2024.

\bibitem{AICity24Paper15}
An~Tuan To, Nam~Minh Tran, Trong-Bao Ho, Thien-Loc Ha, Quang~Tan Nguyen, Chau~Hoang Luong, Thanh-Duy Cao, and Minh-Triet Tran.
\newblock Multi-perspective traffic video description model with fine-grained refinement approach.
\newblock In {\em CVPR Workshop}, Seattle, WA, USA, 2024.

\bibitem{tong2022videomae}
Zhan Tong, Yibing Song, Jue Wang, and Limin Wang.
\newblock Videomae: Masked autoencoders are data-efficient learners for self-supervised video pre-training.
\newblock {\em Advances in neural information processing systems}, 35:10078--10093, 2022.

\bibitem{AICity24Paper12}
Dai~Quoc Tran, Armstrong Aboah, Yuntae Jeon, Maged Shoman, Minsoo Park, and Seunghee Park.
\newblock Low-light image enhancement framework for improved object detection in fisheye lens datasets.
\newblock In {\em CVPR Workshop}, Seattle, WA, USA, 2024.

\bibitem{AICity24Paper10}
Khai~Xuan Trinh, Nguyen~Khoi Nguyen, Bach~Hoang Ngo, Vu~Xuan Dinh, Hung~Minh An, and Vinh Dinh.
\newblock Divide and conquer boosting for enhanced traffic safety description and analysis with large vision language model.
\newblock In {\em CVPR Workshop}, Seattle, WA, USA, 2024.

\bibitem{AICity24Paper43}
Huan Vi and Lap~Quoc Tran.
\newblock Efficient online multi-camera tracking with memory-efficient accumulated appearance features and trajectory validation.
\newblock In {\em CVPR Workshop}, Seattle, WA, USA, 2024.

\bibitem{AICity24Paper36}
Hao~Anh Vo, Sieu Tran, Duc~Minh Nguyen, Thua Nguyen, Tien Do, Duy-Dinh Le, and Thanh~Duc Ngo.
\newblock Robust motorcycle helmet detection in real-world scenarios: Using co-detr and minority class enhancement.
\newblock In {\em CVPR Workshop}, Seattle, WA, USA, 2024.

\bibitem{YoloV7}
Chien-Yao Wang, Alexey Bochkovskiy, and Hong-Yuan~Mark Liao.
\newblock Yolov7: Trainable bag-of-freebies sets new state-of-the-art for real-time object detectors.
\newblock In {\em Proceedings of the IEEE/CVF conference on computer vision and pattern recognition}, pages 7464--7475, 2023.

\bibitem{YOLOR-W6}
Chien-Yao Wang, I-Hau Yeh, and Hong-Yuan~Mark Liao.
\newblock You only learn one representation: Unified network for multiple tasks, 2021.

\bibitem{YOLOR}
Chien-Yao Wang, I-Hau Yeh, and Hong-Yuan~Mark Liao.
\newblock You only learn one representation: Unified network for multiple tasks, 2021.

\bibitem{YOLOv9}
Chien-Yao Wang, I-Hau Yeh, and Hong-Yuan~Mark Liao.
\newblock Yolov9: Learning what you want to learn using programmable gradient information, 2024.

\bibitem{wang2021box}
Hongzhi Wang, Bing Li, Xinyu Xie, Fei Sun, Hua Wang, and Xiaokang Yang.
\newblock Box-grained reranking matching for multi-camera multi-target tracking.
\newblock {\em IEEE Transactions on Pattern Analysis and Machine Intelligence}, 2021.

\bibitem{SR}
Jianyi Wang, Zongsheng Yue, Shangchen Zhou, Kelvin C.~K. Chan, and Chen~Change Loy.
\newblock Exploiting diffusion prior for real-world image super-resolution, 2023.

\bibitem{wang2023videomae}
Limin Wang, Bingkun Huang, Zhiyu Zhao, Zhan Tong, Yinan He, Yi Wang, Yali Wang, and Yu Qiao.
\newblock Videomae v2: Scaling video masked autoencoders with dual masking.
\newblock In {\em Proceedings of the IEEE/CVF Conference on Computer Vision and Pattern Recognition}, pages 14549--14560, 2023.

\bibitem{InternImage}
Wenhai Wang, Jifeng Dai, Zhe Chen, Zhenhang Huang, Zhiqi Li, Xizhou Zhu, Xiaowei Hu, Tong Lu, Lewei Lu, Hongsheng Li, Xiaogang Wang, and Yu Qiao.
\newblock Internimage: Exploring large-scale vision foundation models with deformable convolutions, 2023.

\bibitem{wu2021self}
Yue Wu, Yutian Lin, Yanfeng Wang, Chen Qian, and Yizhou Yu.
\newblock Self-supervised pre-training for transformer-based person re-identification.
\newblock {\em arXiv preprint arXiv:2103.04553}, 2021.

\bibitem{AICity24Paper1}
Zhenyu Xie, Zelin Ni, Wenjie Yang, Yuang Zhang, Yihang Chen, Yang Zhang, and Xiao Ma.
\newblock A robust online multi-camera people tracking system with geometric consistency and state-aware re-id correction.
\newblock In {\em CVPR Workshop}, Seattle, WA, USA, 2024.

\bibitem{yang2023vid2seq}
Antoine Yang, Arsha Nagrani, Paul~Hongsuck Seo, Antoine Miech, Jordi Pont-Tuset, Ivan Laptev, Josef Sivic, and Cordelia Schmid.
\newblock Vid2seq: Large-scale pretraining of a visual language model for dense video captioning, 2023.

\bibitem{AICity24Paper6}
Cheng-Yen Yang, Hsiang-Wei Huang, Pyong-Kun Kim, Zhongyu Jiang, Kwang-Ju Kim, ChungI Huang, Haiqing Du, and Jenq-Neng Hwang.
\newblock An online approach and evaluation method for tracking people across cameras in extremely long video sequence.
\newblock In {\em CVPR Workshop}, Seattle, WA, USA, 2024.

\bibitem{yang2019xlnet}
Zhilin Yang, Zihang Dai, Yiming Yang, Jaime Carbonell, and Quoc~V Le.
\newblock Xlnet: Generalized autoregressive pretraining for language understanding.
\newblock {\em arXiv preprint arXiv:1906.08237}, 2019.

\bibitem{AICity24Paper32}
Ryuto Yoshida, Junichi Okubo, Junichiro Fujii, Masazumi Amakata, and Takayoshi Yamashita.
\newblock Overlap suppression clustering for offline multi-camera people tracking.
\newblock In {\em CVPR Workshop}, Seattle, WA, USA, 2024.

\bibitem{Yu_2020_CVPR}
Fisher Yu, Haofeng Chen, Xin Wang, Wenqi Xian, Yingying Chen, Fangchen Liu, Vashisht Madhavan, and Trevor Darrell.
\newblock Bdd100k: A diverse driving dataset for heterogeneous multitask learning.
\newblock In {\em Proceedings of the IEEE/CVF Conference on Computer Vision and Pattern Recognition (CVPR)}, June 2020.

\bibitem{AICity24Paper37}
Xu Yuehuan and Shuai Jiang.
\newblock Multi-view action recognition for distracted driver behavior localization.
\newblock In {\em CVPR Workshop}, Seattle, WA, USA, 2024.

\bibitem{10.1007/978-3-031-19772-7_29}
Chen-Lin Zhang, Jianxin Wu, and Yin Li.
\newblock Actionformer: Localizing moments of actions with transformers.
\newblock In Shai Avidan, Gabriel Brostow, Moustapha Ciss{\'e}, Giovanni~Maria Farinella, and Tal Hassner, editors, {\em Computer Vision -- ECCV 2022}, pages 492--510, Cham, 2022. Springer Nature Switzerland.

\bibitem{AICity24Paper13}
Hongpu Zhang, Zhe Cui, and Fei Su.
\newblock A coarse-to-fine two-stage helmet detection method for motorcyclists.
\newblock In {\em CVPR Workshop}, Seattle, WA, USA, 2024.

\bibitem{Dino}
Hao Zhang, Feng Li, Shilong Liu, Lei Zhang, Hang Su, Jun Zhu, Lionel~M. Ni, and Heung-Yeung Shum.
\newblock Dino: Detr with improved denoising anchor boxes for end-to-end object detection, 2022.

\bibitem{AICity24Paper24}
Tiantian Zhang, Qingtian Wang, Xiaodong Dong, Wenqing Yu, Hao Sun, Xuyang Zhou, Aigong Zhen, Shun Cui, DONG WU, and He Zhongjiang.
\newblock Augmented self-mask attention transformer for naturalistic driving action recognition.
\newblock In {\em CVPR Workshop}, Seattle, WA, USA, 2024.

\bibitem{zhong2019unsupervised}
Zhun Zhong, Liang Zheng, Donglin Zhang, Deng Cao, and Shuai Yang.
\newblock Unsupervised pre-training for person re-identification.
\newblock In {\em Proceedings of the IEEE Conference on Computer Vision and Pattern Recognition}, pages 8256--8265, 2019.

\bibitem{zhou2020omni}
Yixiao Zhou, Xiaoxiao Liu, Mingsheng Long, Jianmin Zhang, and Trevor Darrell.
\newblock Omni-scale feature learning for person re-identification.
\newblock {\em CVPR}, 2020.

\bibitem{Cycle-GAN}
Jun-Yan Zhu, Taesung Park, Phillip Isola, and Alexei~A. Efros.
\newblock Unpaired image-to-image translation using cycle-consistent adversarial networks.
\newblock In {\em 2017 IEEE International Conference on Computer Vision (ICCV)}. IEEE, Oct. 2017.

\bibitem{Co-DETR}
Zhuofan Zong, Guanglu Song, and Yu Liu.
\newblock Detrs with collaborative hybrid assignments training, 2022.

\bibitem{zong2023detrs}
Zhuofan Zong, Guanglu Song, and Yu Liu.
\newblock Detrs with collaborative hybrid assignments training.
\newblock In {\em Proceedings of the IEEE/CVF international conference on computer vision}, pages 6748--6758, 2023.

\end{thebibliography}
